\documentclass{llncs}
\usepackage{makeidx}  
\usepackage{url}
\usepackage{CJKutf8}
\usepackage{latexsym,graphicx}
\begin{document}
\frontmatter          
\pagestyle{headings}  
\addtocmark{Hamiltonian Mechanics} 
\title{Aicyber's System for NLPCC 2017 Shared Task 2: Voting of Baselines}
\titlerunning{Hamiltonian Mechanics}  
%
\author{Steven Du \and Zhang Xi}

%
%
%
\institute{No. 39 Nanjing Road, Capitaland International Trade Center \\
	 Block B 1604 Tianjin China\\
\email{steven@aicyber.com} \\ \texttt{http://www.aicyber.com}
}

\maketitle              

\begin{abstract}
This paper presents Aicyber's system for NLPCC 2017 shared task 2. It is formed by a voting of three deep learning based system trained on character-enhanced word vectors and a well known bag-of-word model.

\end{abstract}

\section{Introduction}
The NLPCC shared task 2~\cite{qiu2017overview} evaluates the automatic classification techniques for very short texts, the Chinese news headlines. Participants are challenged to identify the category of given texts among 18 classes. The size of training, development and test are 156000, 36000, 36000. The classes in training set are roughly balanced and are equally distributed in development and test set.
The evaluation metrics are macro-averaged precision, recall and F1 score as stated in task guideline~\footnote{\url{http://tcci.ccf.org.cn/conference/2017/dldoc/taskgline02.pdf}}.

This paper describes the second best system submitted by team Aicyber with classification accuracy of 0.825. First, a system overview will be given, then each module will be introduced in detail.


\section{The Aicyber's System}

The submitted system is a voting of three official baseline systems (NBoW, CNN and LSTM ) and a bag-of-word based SVM system. 

Each baseline system's prediction is a voting of that system trained on 5 different word vectors. The sub-systems' architecture, experimental setup, training and development results will be introduced accordingly in the following session. 

\subsection{The Official Baseline Systems }
Three deep learning systems are implemented and released as open source project~\footnote{\url{https://github.com/FudanNLP/nlpcc2017_news_headline_categorization}} by organizer. They are: neural bag-of-words (NBoW) model~\cite{Kalchbrenner2014A,Iyyer2015Deep}, convolutional neural networks (CNN)~\cite{Kim2014Convolutional} and Long short-term memory network~\cite{Hochreiter1997Long}. Hands-on instructions are given to guide participants to reproduce and enhance the baseline systems. The accuracy reported in~\cite{qiu2017overview} are 0.783, 0.763 and 0.747 respectively.

The NBoW model takes an average of the word vectors in the input text and performs classification with a logistic regression layer. It is simple and computationally less expensive than CNN and LSTM system. 


Not like NBoW model who doesn't take the word order into account. The CNN and LSTM (and RNNs) model capture rich compositional information, and have achieved impressive performance in multiple benchmarks~\cite{Kim2014Convolutional,Sheikh2016Learning}. 
~\cite{Zhang2015A} suggested the CNN model need not be complex to realize strong results, as a simple one-layer CNN could achieved state-of-the-art results across several datasets. 
LSTM model has achieved remarkable performance in different sequence learning problems in speech, image and text analysis~\cite{Ghosh2016Contextual,Ji2016Sequential}. It's useful in capturing long-range dependencies in sequences.

The three systems share a similar pipeline for text classification, it takes word/char tokens as input, then tokens will go through word embeddings layers, followed by an average operation (NBoW) or CNN layer or LSTM layer, and a softmax layer at last.

Following sessions will focus on the pre-training of word embeddings layer.

\subsubsection{Word Embeddings}\label{subsection:wordembeddings}

Word embeddings is known as word2vec~\cite{Mikolov2013Distributed}, by default is randomly initialized, for this evaluation pre-trained character and word level embeddings are provided. However we prefer two types of embedding which had superior performance compare with standard approach, these have been verified in dimensional sentiment analysis
task~\cite{Du2017Aicyber}\footnote{\url{https://github.com/StevenLOL/ialp2016_Shared_Task}}.


\paragraph{Character-enhanced Word Embedding}
The first set of word embedding is character-enhanced word embedding~\cite{Chen2015Joint} (CWE). Their study shows semantic meaning of a word is related to its composing characters. Two type of embeddings in CWE, the position-based character embeddings (CWE+P) and cluster-based character embeddings (CWE+L) are used. 

They are trained with window size of 5 and 11, 5 iterations, 5 negative examples, minimum word count of 5, Skip-Gram with starting learning rate of 0.025 , the learned word vectors are of 300 dimensions.

\paragraph{FastText Embedding}
The second set of word embedding is FastText~\cite{Bojanowski2016Enriching}~\footnote{\url{https://github.com/facebookresearch/fastText}}, the idea is to enriching word vectors with sub-word information. Eg, for English, a word vector is associated to its character n-grams.

FastText word embedding is trained with similar setting as CWE training. Please noticed that default minimum character n-gram is 1 for Chinese.





\paragraph{Data Usage for Embedding Training}

Following public available data-sets are used in unsupervised learning of word embeddings:

\begin{enumerate}
	\item Chinese Wikipedia Dumps (Time stamp: 2011-02-
	05T03:58:02Z) 
	\item Douban movie review~\footnote{\url{http://www.datatang.com/data/45075}}
	\item Sogou news corpus~\footnote{\url{http://www.sogou.com/labs/resource/list_news.php}} 	
\end{enumerate}

\paragraph{Training and Evaluation}

Above dataset is preprocessed by jieba~\footnote{\url{https://github.com/fxsjy/jieba}}, after filtering, there are 555571 unique tokens left. Embedding training produces six set of word vectors: CWE-L-W5,CWE-L-W11,CWE-P-W5,CWE-P-W11,FastText-W5 and FastText-W11 (W denotes window size).

\begin{CJK}{UTF8}{gbsn}
To verify the correctness of word embedding, we examines the nearest neighbor of a given Chinese word, eg 高兴(happy). The result from FastText-W11 is clearly different from others, 5 single character words appear in the top 10 (0 for other embeddings). This indicates FastText doesn't work properly for Chinese with large window size, in which the character n-grams, especially unigram become overestimate. Thus FastText-W11 is dropped.


\end{CJK}

\subsubsection{Training of Official Baseline systems}
With 5 embedding from above, 15 (5*3) systems are formed. We use default setting for NBoW, LSTM system. For the CNN system, only one convolution layer is used (filter size is 3).

System is trained only on the 156000 training data, and evaluated on 36000 development data, we use accuracy as performance metrics.



\subsubsection{Result and Discussion}


The results of official baselines are presented in Table~\ref{tb:EvalFeatureLSVR}, to make a fair comparison, systems trained on randomized character/word vectors (length is 300) are also included.

\begin{table}[h]
	\small
	\centering
	\begin{tabular}{|l|l|r|}
		\hline \multicolumn{3}{|c|}{\textbf{Official Baseline systems }}   \\ 
		\hline \bf Network Type & \bf Embeddings & \bf Development Accuracy   \\ \hline
		NBoW & Randomized Char & 0.715 \\
		NBoW & Randomized Word  & 0.779    \\
		NBoW & CWE-L-W5 & 0.814    \\ 
		NBoW & CWE-P-W5 & 0.816    \\ 
		NBoW & FastText-W5 & 0.812    \\ 
		NBoW & CWE-L-W11 & 0.816    \\ 
		NBoW & CWE-P-W11 & 0.816  \\ \hline
		CNN & Randomized Char & 0.718    \\
		CNN & Randomized Word & 0.763  \\
		CNN & CWE-L-W5 & 0.822  \\ 
		CNN & CWE-P-W5 & 0.823   \\
		CNN & FastText-W5 & 0.820   \\
		CNN & CWE-L-W11 & \textbf{0.824}    \\
		CNN & CWE-P-W11 & 0.821    \\	\hline
		LSTM & Randomized Char & 0.691    \\
		LSTM & Randomized Word & 0.728    \\ 
		LSTM & CWE-L-W5 & 0.808    \\ 
		LSTM & CWE-P-W5 & 0.805   \\ 
		LSTM & FastText-W5 & 0.801     \\ 
		LSTM & CWE-L-W11 & 0.807    \\ 
		LSTM & CWE-P-W11 & 0.806     \\

		
		\hline
	\end{tabular}
	\caption{\label{tb:EvalFeatureLSVR} The official baseline systems's accuracy on development set. CNN system with CWE-L-W11 is the best system and achieve 0.824.}
\end{table}

It's obvious that systems trained with pre-trained embedding are much better than those with randomized embeddings. System with word embeddings give better result than those use characters embedding. CNN is the most accurate system. For different embedding types, the FastText is under performance the others. Difference between CWE-P and CWE-L is negligible.



\subsection{Bag-of-Word model}

The official released systems are relatively strong. We also seek alternatives to tackle classification problem. Starting with a well known baseline system, the bag-of-word model. It's commonly used in text classification where the occurrence of each word is used as a feature for training classifiers. Support Vector Machine~\cite{Vapnik1995The} (SVM) with linear kernel was considered to be one of the best classifiers~\cite{Forman2003An,Yang1999A}.


This system is trained on 156000 training data, and validated on 36000 development set. Table~\ref{tb:bow} shows BoW model could obtain 0.791 classification accuracy. The result is much better than all deep learning system with randomized embeddings, this finding demonstrate the importance of pre-trained word vectors.




\begin{table}
	\small
	\centering
	\begin{tabular}{|l|l|r|}
		\hline \multicolumn{3}{|c|}{\textbf{BoW Vs Deep-learning sytem with Randomized Word Vector }}   \\ 
		\hline \bf Features & \bf Classifiers & \bf Development Accuracy   \\ \hline
		Bag-of-Word & Linear SVM & \textbf{0.791} \\
		
		\hline
		Randomized Word Vector & LSTM & 0.728 \\
		Randomized Word Vector & CNN & 0.763 \\
		Randomized Word Vector & NBoW & 0.779 \\

		
		
		\hline
	\end{tabular}
	\caption{\label{tb:bow} Classification accuracy of bag-of-word model is better than all deep learning system with randomized embeddings.}
\end{table}

To summarize, the submitted system is an ensemble of three deep learning based systems and a conventional BoW model, it's truly a voting of baselines. The final classification accuracy is 0.826 measured on the development set. 

\section{Discussion and Further Improvement}

Compare BoW method in Table~\ref{tb:bow} with the best single system in Table~\ref{tb:EvalFeatureLSVR}, the difference is only 0.03, we consider that the BoW indeed is well suited for the Chinese headline classification. Because the headline appears to be clear, concise and powerful, the usage of words in headline is precisely selected by professional editors. The importance of words make the BoW works well for this task.

The best single system achieved 0.824 classification accuracy on development set, while the voting system scored 0.826 on same dataset. Voting method provides marginal improvement in this work.

Word embedding training in Section 2.1.4 is kind of unsupervised pre-training, but only limited to the embedding layer. Study in~\cite{Dai2015Semi} shows recurrent language models  and sequence autoencoder could used to pre-train not only the embedding layer but also the LSTM layer. On five benchmarks that they tried, LSTMs can reach or surpass the performance levels of all previous baselines. As shown in Table~\ref{tb:EvalFeatureLSVR}, LSTM didn't beat CNN or NBoW model, using pre-training could boost LSTM's performance.


\section{Conclusion}
In this paper we presented our approaches to tackle Chinese news headline categorization challenge. A voting system consists of three deep-learning based system build on five different embedding layers and a BoW model ranked 2nd among 32 teams. 

\makeatletter
\renewenvironment{thebibliography}[1]
{\section*{\refname}
	\small
	\list{}%
	{\settowidth\labelwidth{}%
		\leftmargin\parindent
		\itemindent=-\parindent
		\labelsep=\z@
		\if@openbib
		\advance\leftmargin\bibindent
		\itemindent -\bibindent
		\listparindent \itemindent
		\parsep \z@
		\fi
		\usecounter{enumiv}%
		\let\p@enumiv\@empty
		\renewcommand\theenumiv{}}%
	\if@openbib
	\renewcommand\newblock{\par}%
	\else
	\renewcommand\newblock{\hskip .11em \@plus.33em \@minus.07em}%
	\fi
	\sloppy\clubpenalty4000\widowpenalty4000%
	\sfcode`\.=\@m}
{\def\@noitemerr
	{\@latex@warning{Empty `thebibliography' environment}}%
	\endlist}
\def\@cite#1{#1}%
\def\@lbibitem[#1]#2{\item[]\if@filesw
	{\def\protect##1{\string ##1\space}\immediate
		\write\@auxout{\string\bibcite{#2}{#1}}}\fi\ignorespaces}
\makeatother

%
%

\end{document}